\documentclass[10pt,twocolumn,letterpaper]{article}

\usepackage[pagenumbers]{cvpr} 

\usepackage{graphicx}
\usepackage{amsmath}
\usepackage{amssymb}
\usepackage{booktabs}
\usepackage{multirow}

\usepackage[pagebackref,breaklinks,colorlinks]{hyperref}

\usepackage[capitalize]{cleveref}
\crefname{section}{Sec.}{Secs.}
\Crefname{section}{Section}{Sections}
\Crefname{table}{Table}{Tables}
\crefname{table}{Tab.}{Tabs.}

\def\cvprPaperID{2505} 
\def\confName{CVPR}
\def\confYear{2023}

\begin{document}

\title{MIAD: A Maintenance Inspection Dataset for Unsupervised Anomaly Detection}


\author{
Tianpeng Bao\textsuperscript{1} \quad
Jiadong Chen\textsuperscript{1} \quad
Wei Li\textsuperscript{1} \quad
Xiang Wang\textsuperscript{1} \quad
Jingjing Fei\textsuperscript{1} \quad
Liwei Wu\textsuperscript{1} \quad\\
Rui Zhao\textsuperscript{1,3} \quad
Ye Zheng\textsuperscript{2} \quad\\
SenseTime Research\textsuperscript{1} \quad
JD.com, Inc.\textsuperscript{2}\\
Qing Yuan Research Institute, Shanghai Jiao Tong University, Shanghai, China\textsuperscript{3}\\
{\tt\small \{baotianpeng, feijingjing1, wangxiang, liwei1, zhaorui, wuliwei\}@sensetime.com}
\\{\tt\small chenjiadong1@senseauto.com} \quad {\tt\small zhengye12@jd.com}
}

\maketitle

\begin{abstract}


Visual anomaly detection plays a crucial role in not only manufacturing inspection to find defects of products during manufacturing processes, but also maintenance inspection to keep equipment in optimum working condition particularly outdoors. Due to the scarcity of the defective samples, unsupervised anomaly detection has attracted great attention in recent years. However, existing datasets for unsupervised anomaly detection are biased towards manufacturing inspection, not considering maintenance inspection which is usually conducted under outdoor uncontrolled environment such as varying camera viewpoints, messy background and degradation of object surface after long-term working. We focus on outdoor maintenance inspection and contribute a comprehensive Maintenance Inspection Anomaly Detection (MIAD) dataset which contains more than 100K high-resolution color images in various outdoor industrial scenarios. This dataset is generated by a 3D graphics software and covers both surface and logical anomalies with pixel-precise ground truth. Extensive evaluations of representative algorithms for unsupervised anomaly detection are conducted, and we expect MIAD and corresponding experimental results can inspire research community in outdoor unsupervised anomaly detection tasks. Worthwhile and related future work can be spawned from our new dataset.

\end{abstract}
\section{Introduction}
Anomaly detection plays a crucial role in not only manufacturing inspection but also maintenance inspection.
Manufacturing inspection is intended to find defect of products during manufacturing process in many industrial fields, such as electronics~\cite{review_aoi}, metals~\cite{artical_metallic_surface}, fabrics~\cite{aitex} and food~\cite{review_food}.
Maintenance inspection is intended to find whether equipment or product is in optimum working condition after leaving the factory particularly outdoors, such as inspection of power transmission lines~\cite{review_power_transmission}, photovoltaic power station~\cite{artical_photovoltaic}, wind power plant~\cite{review_wind_turbine} and overhead catenary system~\cite{artical_catenary}.
Among various monitoring approaches including ultrasonic and X-ray, optical inspection is the most basic, low-cost and frequently used type and is considered non-contact and non-destructive~\cite{review_aoi, review_wind_turbine}.

\begin{figure}[t]
    \centering
    \includegraphics[width=0.45 \textwidth]{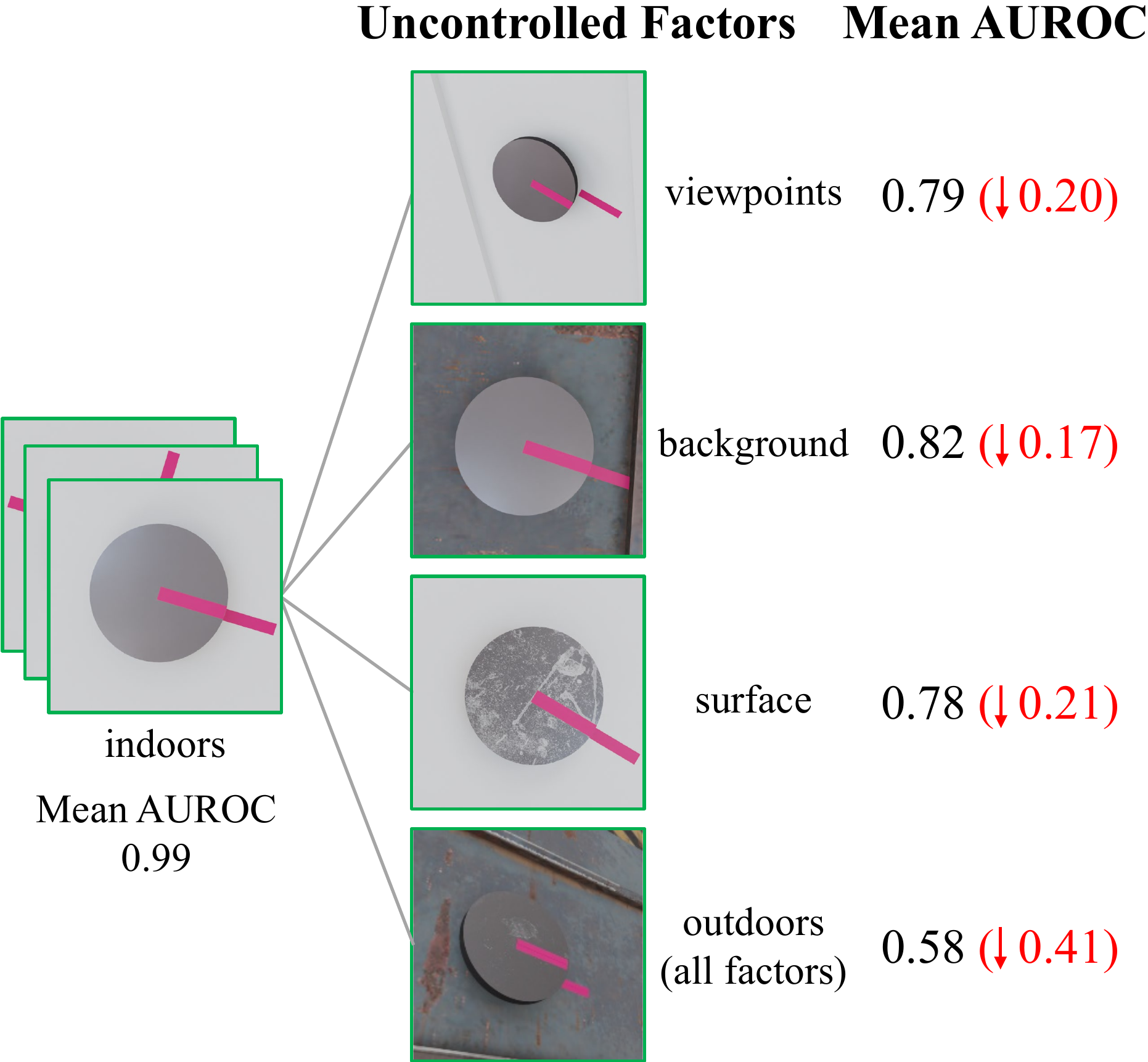}
    \vspace{-5pt}
    \caption{Images for maintenance inspection are usually captured outdoors and suffer from varying camera viewpoints, messy background and degradation of object surface after working for a long time, which makes the distribution of non-defective samples more complicated.
    Some state-of-the-art methods can achieve AUROC of 0.99 on the toy dataset for witness marks with indoor setting. 
    But when we vary each outdoor uncontrolled factor, we observe obvious drops in performance.
    We show the mean AUROC across state-of-the-art methods on MVTec AD including PatchCore~\cite{patchcore}, Reverse Distillation~\cite{rd}, FastFlow~\cite{fastflow} and DRAEM~\cite{zavrtanik2021draem}.
    The detailed metrics are shown in Table~\ref{tab:exp_uncontrolled_factors_wm}.}
    \label{fig:3factors}
    \vspace{-15pt}
\end{figure}

Due to the scarcity of defective samples and abundance of non-defective samples,
research community as well as industrial community has paid increased attention to unsupervised setting of anomaly detection, i.e. algorithms should be trained solely on non-defective images and be tested on both non-defective and diverse defective images.

The problem of unsupervised anomaly detection for manufacturing inspection~\cite{survey_uad} has been thorough researched, which benefits from plenty of emerging datasets especially MVTec AD~\cite{mvtec_ad}.
MVTec AD is the first comprehensive dataset for unsupervised anomaly detection in industrial inspection which provides pixel-accurate ground truth regions and allows to evaluate algorithms at both image level and pixel level.
Since MVTec AD is collected indoors under controlled environments, images of the same category are roughly aligned at pixel level with a clean background.
Based on this prior knowledge, 
PatchCore~\cite{patchcore} achieves state-of-the-art performance with image-level AUROC of more than 0.99.
However, when we try to apply PatchCore and other representative methods to maintenance inspection in the outdoors, we observe significant drops in performance.
Besides, there are only hundreds of images in the training dataset per category, which is enough for indoor scenarios but is not sufficient for outdoor scenarios due to the complex distribution of non-defective samples. 
In a word, existing datasets including MVTec AD are not suitable for research on outdoor maintenance inspection due to simple pixel-level alignment and limited training dataset size.

The imaging equipment for maintenance inspection are usually carried outdoors by unmanned aerial vehicles (UAV) or track inspection vehicles~\cite{artical_catenary} 
under uncontrolled environment.
As shown in Figure~\ref{fig:3factors}, images for outdoor maintenance inspection suffer from varying camera viewpoints, messy background and the degradation of object surface (e.g. dust or rust), and it is reasonable to doubt whether these uncontrolled factors break the prior assumption about pixel-level alignment.
In order to verify this doubt, we synthesize a toy dataset of witness mark by control variable technology.
Experiments on this toy dataset show that uncontrolled viewpoints, background, and surface individually have a negative effect on the accuracy of the state-of-the-art algorithms, and the combination of these uncontrolled factors results in a more challenging task.

Therefore, we build and release a maintenance inspection dataset,  named MIAD\footnote{\url{https://miad-2022.github.io/}}, for unsupervised anomaly detection.
Considering the accessibility of various industrial scenarios and high-cost labeling of pixel-precise ground truth, we make use of BlenderProc~\cite{denninger2019blenderproc}, a 3D graphics software, to build the 3D scenes and automatically generate 2D color images with pixel-precise ground truth.
The dataset mimics seven outdoor scenarios including photovoltaic module, wind turbine blade, nut and bolt, which is inspired by real-world maintenance inspection.
Moreover, the area under the receiver operating characteristic curve (AUROC) is utilized as a image-level metric by existing datastes, which treats the performance at the high false positive rate (FPR) and the low FPR equally.
We adopt an additional metric to focus on the performance at the low false alarm rate which is pursued by industrial community.

In summary, we make the following contributions:
\begin{itemize}
    \item We introduce a novel maintenance inspection dataset for unsupervised anomaly detection under uncontrolled environments.
    It consists of more than 100K high-resolution color images with pixel-precise ground truth, an order of magnitude larger than existing datasets.
    Moreover, it covers both surface anomalies (such as scratches, and dents occurring on the surface of objects) and logical anomalies (incorrect relative position of two or more parts, such as missing one part, or looseness between the nut and bolt).
    \item We conduct extensive evaluations on MIAD with representative algorithms for unsupervised anomaly detection.
    We also introduce an additional metric that focuses on the image-level performance at the low FPR.
    Experiments show that the evaluated methods do not perform equally well across surface and logical anomalies. There is a long way to go in applying unsupervised anomaly detection to outdoor maintenance inspection.
\end{itemize}
\section{Related Work}\label{sec:related_work}

The progress of research in industrial anomaly detection has been strongly driven by the development of datasets.
Here, We briefly review related datasets and methods for unsupervised anomaly detection.

\subsection{Datasets for Unsupervised Anomaly Detection}
\subsubsection{Datasets in Manufacturing Inspection}

\begin{table*}
  \centering
  \resizebox{1.0\linewidth}{!}{
  \begin{tabular}{cccccccc}
    \toprule
   \textbf{Datasets} & \textbf{\#Category} & \textbf{\#Defect types} & \textbf{\#Train} & \textbf{\#Test(good+defective)} & \textbf{Scene \& Env} & \textbf{Real/Synthetic} \\
    \midrule
    AITEX & 1 & 12 & 0 & 140+105  & Manufacturing \& Indoor & Real \\
    BTAD & 3  & 3 & 1799 & 1031(total) & Manufacturing \&Indoor & Real \\
    DAGM & 10 & 10 & 6900 & 4600(total)  & Manufacturing \&Indoor & Synthetic \\
    ELPV & 2 & 2 & 1968 & 656(total) & Manufacturing \& Indoor & Real \\
    KolektorSDD &1 &  1& 133 &266(total) & Manufacturing \& Indoor & Real \\
    MTD &  2 & 8 & 448  & 896(total)& Manufacturing \& Indoor & Real \\
    MVTec AD & 15 &  73  & 3629 & 467+1258 & Manufacturing \& Indoor & Real \\
    MVTec 3D AD & 10 & 41  & 2656 & 248+948 & Manufacturing \& Indoor & Real \\
    MVTec LOCO AD & 5 & 89 & 1772 & 575+993  & Manufacturing \& Indoor & Real \\
    \midrule
    CPLID & 1 & 1 &  1186 & 770(total) & Maintenance \& Outdoor & Real \& Synthetic \\
    \textbf{MIAD} & 7 & 13 & \textbf{70000} & \textbf{17500+17500} & Maintenance \& Outdoor & Synthetic \\
    \bottomrule
  \end{tabular}}
  \caption{Comparison of datasets for unsupervised anomaly detection. 
  Our MIAD dataset focus on outdoor maintenance inspection and is an order of magnitude larger than existing datasets.
  }
  \label{tab:compare_datasets}
  \vspace{-10pt}
\end{table*}

\textbf{AITEX}~\cite{aitex} is a public dataset for textile inspection, which consists of 245 images of different fabric textures obtained from a real production plant. \textbf{BTAD}~\cite{mishra2021vt} contains a total of 2830 real-world images of 3 industrial products showcasing body and surface defects. \textbf{DAGM}~\cite{dagm} is a synthetic dataset for defect detection on textured surfaces. It consists of multiple categories generated by a different texture model and defect model, each consisting of 1000 normal samples and abnormal images with annotated defect. \textbf{ELPV}~\cite{Deitsch2019elpv} is collected based on electroluminescence (EL) imaging of polycrystalline photovoltaic (PV) modules. It contains 2624 8-bit grayscale images of functional and defective solar cells. All images are the 300x300 pixels with varying degree of degradations extracted from 44 different solar modules. \textbf{KolektorSDD}~\cite{Tabernik2019JIM} is proposed for surface-defect detection for an industrial semi-finished product where the number of defective items available for the training is limited. \textbf{MTD}~\cite{huang2020surface} contains 1344 images, with the cropped ROIs of 6 kind of magnetic tile and pixel-level labels. To simulate the manufacturing process in real assembly line, this dataset collects images under multiple illumination conditions for each given magnetic tile.
\textbf{MVTec AD}~\cite{mvtec_ad} is a widely used unsupervised anomaly detection dataset, which contains 5354 high-resolution normal and anomalous images of 15 different real-world products. Most of recent unsupervised anomaly detection methods are driven from this dataset. \textbf{MVTec 3D AD}~\cite{bergmann2021mvtec} is proposed recently to encourage research into 3D anomaly detection and segmentation.
It contains over 4000 high-resolution 3D scans of industrially manufactured products across 10 categories. Each sample is represented by an organized point cloud and a corresponding RGB image with a one-to-one mapping between the pixels in the point cloud and those in the RGB image.
\textbf{MVTec LOCO AD} (MVTec Logical Constraints Anomaly Detection)~\cite{mvtec_loco}  includes both structural and logical anomalies with 3644 images from five different categories inspired by real-world industrial inspection scenarios.

The details of aforementioned datasets are summarized in Table~\ref{tab:compare_datasets}.
Compared with our MIAD dataset, these datasets are all collected indoors and does not take into consideration uncontrolled environment such as random camera viewpoints, messy background, degradation of object surface (rust or dust), etc. In addition, their datasets have small scales, so it may unable to fully exploit the potential of unsupervised anomaly detection algorithms and to adequately validate them.

\subsubsection{Datasets in Maintenance Inspection}
Most of the above datasets are collected from indoor scenarios under controlled environments. We also survey some outdoor datasets which are rarely mentioned by the unsupervised anomaly detection community.

Xiaoxia Li et al.~\cite{artical_photovoltaic} collected two typical PV module visible defects: snail trail and dust shading by the UAV inspection system.
The railway catenary dataset used in~\cite{artical_catenary} consists of the catenary support device images captured from an approximately 100-km line along a high-speed railway, in which 2000 catenary support devices and 40000 fasteners exist.
\textbf{CPLID}~\cite{tao2018detection} is proposed to detect the defect of grid insulator, which contains 1956 high resolution images. It is captured by an UAV with a DJI M200 camera at a resolution of 4608 × 3456 pixels and stored in BMP format. 

However, the PV module~\cite{artical_photovoltaic} and the railway catenary\cite{artical_catenary} datasets for outdoor inspection are not publicly available, and thus they cannot be used to verify any proposed results or make comparisons between different methods. Most of negative images in CPLID~\cite{tao2018detection} are synthesized by data augmentation, making this dataset has insufficient number of valid outlier samples.
Moreover, these datasets only focus on one specific industry and cannot be utilized to analyze the generalization ability of algorithms.
In this paper, we present an outdoor maintenance inspection dataset with more than 100K annotated images and covering various outdoor industrial maintenance inspection scenarios including photovoltaic panels and overhead catenary.

\subsection{Methods for Unsupervised Anomaly Detection}


Following the survey~\cite{survey_uad}, we classify the unsupervised anomaly detection algorithms into four classes: reconstruction-based, representation-based, normalizing flow-based and data augmentation-based methods. 
We will select one or two representative methods from each class to make an analysis on our MIAD dataset in Section \ref{sec:exp}.

\subsubsection{Reconstruction-based Methods}

Reconstruction-based methods are designed to score anomaly by reconstruction error.
A neural network is trained only on the normal images to generate high-fidelity input reconstructions and, it is not able to reconstruct abnormal images correctly in inference.
The greater probability of anomaly there is, the higher anomaly scores of a pixel will be.
Typical algorithms mainly include auto-encoders~\cite{lecun1989,structuring_ae,ssim_ae}, variational auto-encoders~\cite{vae} and generative adversarial networks~\cite{gan}.
Recently the teacher-student framework~\cite{stpm, rstpm, rd} is  adopted for unsupervised anomaly detection.
In the training phase, only the student network is trained to reconstruct the multi-layer feature of the teacher network whose frozen parameters are pre-trained on the ImageNet.
In the testing phase, both teacher and student networks are fed with a sample and the corresponding anomaly map is computed based on the difference of the two multi-layer features.
Especially, the student network of Reverse Distillation~\cite{rd} takes teacher model’s one-class embedding rather than raw images as input to reverse data flow in knowledge distillation, which achieves promising results on the MVTec AD benchmark due to heterogeneity of the teacher and student networks. UniAD~\cite{uni_ad} accomplishes anomaly detection for multiple classes with a unified framework by designing improvements base on the transformer.

\subsubsection{Representation-based Methods}
Representation-base methods extract discriminative features from normal images and further build the normal distribution. Anomaly score results are  obtained by measuring the distance between the test images with the distribution of normal images.
Typical methods include SPADE~\cite{spade}, PaDiM~\cite{padim}, FYD~\cite{zheng2022focus},  PatchCore~\cite{patchcore}, etc.
However, these methods are CPU/GPU memory-consuming when there are thousands of training samples.
Their training time complexities scale linearly with the dataset size, and require large memory allocation for gallery features.

%
CFA~\cite{cfa} consists of a learnable patch descriptor to learn embed representations and a scalable memory bank independent of the size of the dataset. It adopts transfer learning to increase the normal feature density so that abnormal features can be clearly distinguished by applying patch descriptor and memory bank to a pre-trained CNN.

\subsubsection{Normalizing Flow-based Methods}

Normalizing Flows~\cite{nflow} are intended to learn transformations between data distributions and well-defined standard normal distribution.
Recently, some work began to use it for unsupervised
anomaly detection and localization. In these methods, normal image features are embedded into standard normal distribution and the probability is used to identify and locate anomalies. DifferNet~\cite{differnet} achieved good image level
anomaly detection performance by using it to estimate the
precise likelihood of test images. Unfortunately, this work failed to obtain the exact anomaly localization results since they flattened the outputs of feature extractor. 
CFLOW-AD~\cite{cflow} proposes to use hard code position embedding to leverage the distribution learned by Normalizing Flows, which probably under performs at more complicated datasets.
Furthermore, FastFlow~\cite{fastflow} implements efficient 2D normalizing flows and use it as the anomaly estimator, which can learns to transform  the visual input as a tractable distribution at train stage and obtain the probability to identify anomalies at inference stage.

\subsubsection{Data Augmentation-based Methods}

The core idea of data augmentation-based methods is to simulate the anomalies through augmentation, which is data-dependent and hand-crafted.
DRAEM~\cite{zavrtanik2021draem} learns a joint representation of an anomalous image and its anomaly-free reconstruction, while simultaneously learning a decision boundary between normal and anomalous examples.
CutPaste~\cite{li2021cutpaste} uses a simple data augmentation strategy that cuts an image patch and pastes at a random location of a large image to train the feature extraction model.
\section{Benchmark}

In this section, we will first describe the details of our MIAD dataset.
And then we will define the metrics for anomaly classification and segmentation performance on MIAD to build a benchmark.

\subsection{Dataset Description}

\begin{figure*}[t]
    \centering
    \includegraphics[width=1.0 \textwidth]{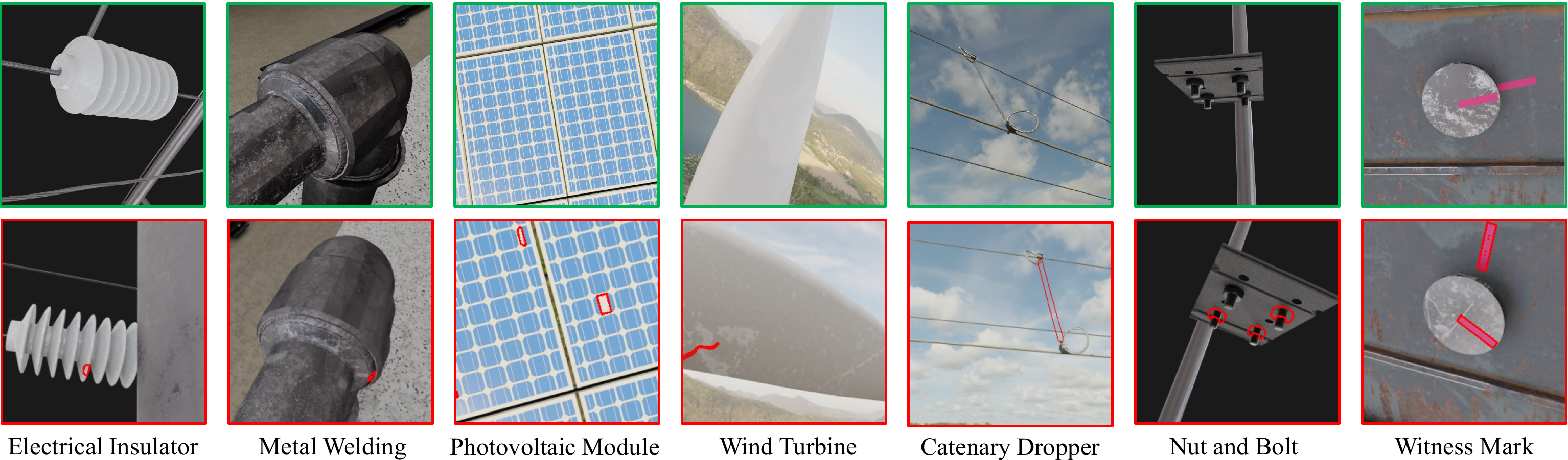}
    \vspace{-15pt}
    \caption{Example images of the MIAD dataset for seven maintenance inspection scenarios.
    The test set for each scenario comprises non-defective (top row) and defective images (bottom row). The first four scenarios contain surface anomalies, and the rest contain logical anomalies.
    Pixel-precise annotations are provided for all anomalies.
}
    \label{fig:dataset_samples}
    \vspace{-5pt}
\end{figure*}

\begin{table*}
  \centering
  \begin{tabular}{cccccc}
    \toprule
\textbf{Anomaly} & \textbf{Scenarios} & \textbf{Uncontrolled Factors} & \textbf{Day/Night} & \textbf{\#Defect Types} & \textbf{\#Defect Pixel Ratio} \\
    \midrule
Surface & Electrical Insulator & UV, UB, US & Night & 1 & 0.04\% \\
    & Metal Welding & UV, US & Day & 2 & 0.10\% \\
    & Photovoltaic Module & UV & Day & 3 & 0.11\% \\
    & Wind Turbine & UV, UB, US & Day & 1 & 0.05\% \\
    \hline
Logical & Catenary Dropper & UV, UB & Day & 3 & 0.37\% \\
    & Nut and Bolt & UV, UB & Night & 3 & 0.36\% \\
    & Witness Mark & UV, UB, US & Day & 1 & 0.70\% \\
    \bottomrule
  \end{tabular}
  \caption{Overview of the MIAD dataset.
  In the third column, UV, UB and US stand for Uncontrolled Viewpoints, Uncontrolled Background and Uncontrolled Surface, respectively.
  The defect pixel ratio equals the number of defective pixels divided by the number of all pixels. }
  \label{tab:statistic}
  \vspace{-10pt}
\end{table*}

The MIAD dataset consists of seven outdoor maintenance inspection scenarios.
All scenarios are inspired by real-world maintenance inspection including metal welding for oil transmission pipeline,
photovoltaic modules in photovoltaic power station~\cite{artical_photovoltaic}, wind turbine blades in wind power plant~\cite{review_wind_turbine}, the catenary dropper in overhead catenary system~\cite{artical_catenary}.
The electrical insulator, nut and bolt, and witness mark are widely used in many industrial fields such as power transmission lines~\cite{review_power_transmission} and catenary support devices~\cite{artical_catenary}.
Note that outdoor maintenance inspection is usually conducted during the day, but in some special cases it has to be conducted during the night.
For example, the track inspection vehicles for catenary support device only work at night since the railway is occupied by transporting passengers during the day.
%
The example images of each scenarios are illustrated in Figure~\ref{fig:dataset_samples}.

There are a total of 105000 color images with 512 $\times$ 512 pixel high-resolution.
The training set comprises 7000 images without any defect, 10000 of each scenarios.
The test set comprises 35000 images, 2500 non-defective and 2500 defective images of each scenarios.
Further information about each category is summarized in Table \ref{tab:statistic}.

Different from the existing datasets for manufacturing inspection which mainly collected in controlled environment, the MIAD dataset focus on the impact of uncontrolled environment for maintenance inspection.
Images acquisition in maintenance inspection is usually conducted by a UAV system, which results in varying camera viewpoints (uncontrolled viewpoints).
Messy background (uncontrolled background) is also inevitable when imaging in an open environment.
Moreover, degradation of object surface (uncontrolled surface) after long-term outdoor working makes the distribution of non-defective objects more complex.
For convenience, we utilize UV, UB, US to denote uncontrolled viewpoints, uncontrolled background and uncontrolled surface in this paper, respectively.

\begin{figure*}[t]
    \centering
    \vspace{-10pt}
    \includegraphics[width=1.0 \textwidth]{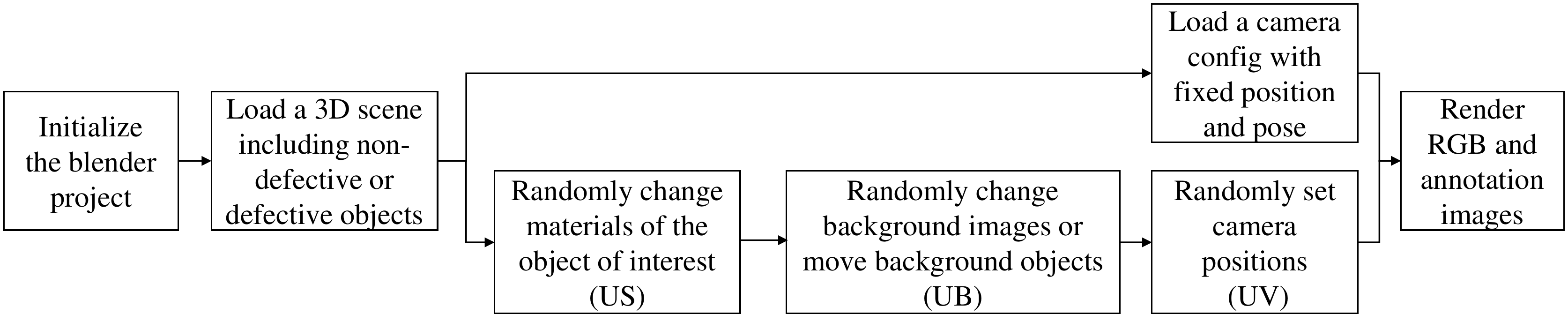}
    \vspace{-10pt}
    \caption{
    The basic pipeline (top flow) and a modified pipeline (bottom flow) for rendering photorealistic data by BlenderProc.
    }
    \label{fig:pipeline}
    \vspace{-10pt}
\end{figure*}

We utilize BlenderProc\cite{denninger2019blenderproc}, a procedural Blender pipeline for photorealistic rendering, to generate our MIAD dataset.
A basic pipeline for rendering with BlenderProc is shown as the top flow in Fig~\ref{fig:pipeline}.
After some basic initialization of the blender project (e.g. configures computing device, creates a camera), a 3D scene is loaded and a camera is set with a predefined position and pose inside this scene.
Then the RGB and semantic segmentation images can be rendered simultaneously based on the physical based rendering~\cite{PBR} technology.
In order to simulate the uncontrolled surface, background and viewpoints in maintenance inspection, three modules are implemented.
The bottom flow in Fig~\ref{fig:pipeline} depicts the modified pipeline and the details will be describes in the following three subsections.



\subsubsection{Uncontrolled Surface}
In the outdoors, the surface of objects may deteriorate (such as dust or rust) due to bad weather or other environment, resulting in random irregular textures.
To simulate the uncontrolled surface, we randomly change materials of the object of interest by a mix shader which can mix the original shader with another one like dust.
The original and additional materials can been found in blenderco textures library\footnote{\url{https://blenderco.cn/category/tiet}}. 
Plenty of similar texture files are downloads and they are randomly split into training set and test set to prevent the data leakage,

\subsubsection{Uncontrolled Background}

To simulate the uncontrolled background, we randomly change the background images when the scenario (e.g. the wind turbine in our MIAD dataset) is during the day. 
Since the real-world images for maintenance inspection at night fill with regular black background like~\cite{artical_catenary}, we randomly move background objects to simulate the messy background when the scenario (e.g. the nut and bolt in our MIAD dataset) is during the night.


\subsubsection{Uncontrolled Viewpoints}

The images varies with the uncontrolled camera viewpoints which has 6 degree of freedom.
In order to capture the object of interest, the 3D pose is constrained based on that the camera always look at a point of interest on the object, while the 3D position can been randomly set.
For convenient, we sample the camera position in spherical coordinate rather than Cartesian coordinate.
In other word, the camera position can be denoted as $(r, \theta, \phi)$, where $r$ is the radius, $\theta$ is the azimuth angle and $\phi$ is the elevation angle.
$r$ affects the size of objects, while $\theta$ and $\phi$ jointly affects the visible region of objects.
The sampling interval of $r$ is shared by training and test set to make the dataset follow similar distribution.
To prevent the data leakage, the $\theta$ and $\phi$ for training images satisfy
\begin{equation}
\theta, \phi \in \cup_{n=0}^{179}~[2n, 2n+1),
\end{equation}
and
the $\theta$ and $\phi$ for test images satisfy
\begin{equation}
\theta, \phi \in \cup_{n=0}^{179}~[2n+1, 2n+2),
\end{equation}
where $\theta$ and $\phi$ are both in degrees.

\subsection{Evaluation Metrics}

All evaluated algorithms should provide a one-channel anomaly map, in which large values indicate that a certain pixel belongs to an anomalous region.
And the maximum score over all pixels in a given anomaly map is regarded as the image-level anomaly score.

To assess the image-level anomaly classification performance, we adopt AUROC as a evaluation metric following MVTec AD~\cite{mvtec_ad}.
Although AUROC is a independent of thresholds, a threshold must be determined to make a binary decision when algorithms are applied in industrial scenarios.
Moreover, AUROC treats the performance at the high FPR and the low FPR equally, but industrial community pursues better performance only at the low FPR because lower false alarm rate brings lower cost for manually checking the report from algorithms.
FPR is usually required no larger than 1\%, so we propose to use recall at 1\% FPR as a additional image-level metric to measure whether a method is applicable in industry.
The determined threshold is calculated according to 1\% FPR on testset.


In order to assess the anomaly segmentation capability, a naive way is to calculate the pixel-level AUROC by regarding the classification result of each pixel as a sample. 
However, the defect pixel ratio of all scenarios in Table~\ref{tab:statistic} is less than 0.01 and this means pixel-level AUROC can easily surpass 0.99 if all pixels are predicted as non-defective, which indicates that pixel-level AUROC is not a proper metric to reflect different performance of evaluated methods, especially in outdoor inspection.
Following MVTec AD~\cite{mvtec_ad}, a normalized per-region overlap (PRO) between segmentation and ground truth is calculated and the area under the PRO curve (AUPRO) is adopted as a pixel-level metric.
For the convenience of comparing the performance on MVTec AD and MIAD of the same methods, the normalized area under the PRO curve is up to pixel-level FPR of 30\%.

\section{Experiments}\label{sec:exp}

In this section, we conduct a thorough evaluation of multiple state-of-the-art methods for unsupervised anomaly detection on our dataset to serve as a baseline for future methods. 
The strengths and weaknesses of each method are discussed on the various surface and logical anomalies.
Moreover, we analysis the impact of three uncontrolled factors of outdoors and the impact of training dataset size in a quantitative way.

\subsection{Evaluated Methods}

We select Reverse Distillation~\cite{rd}, PatchCore~\cite{patchcore}, FastFlow~\cite{fastflow} and DRAEM~\cite{zavrtanik2021draem} as representative of reconstruction-, representation-, normalizing flow- and data augmentation-based methods, respectively.
The naive method L2 Auto-Encoder as described by ~\cite{ssim_ae} is also adopted as a baseline.
Since the above methods train separate models for different scenarios,  we also adopt UniAD~\cite{uni_ad} which shares the same parameters  for multiple scenarios.
%
Detailed information of each algorithms including the input size, data augmentation and neural network can be found in Appendix.

\begin{table}
  \centering
  \resizebox{\linewidth}{!}{
    \begin{tabular}{ccccccc}
        \toprule
    \textbf{Scenarios} &  \textbf{RD} & \textbf{PatchCore} & \textbf{FastFlow} & \textbf{DRAEM} & \textbf{AE} & \textbf{UniAD}\\
        \midrule
    Electrical & 0.68 & 0.55 & 0.54 & \textbf{0.81} & 0.49 & 0.50\\
     Insulator & 0.01 & 0.01 & 0.01 & \textbf{0.24} & 0.01 & 0.01\\
        \midrule
    Metal      & 0.91 & \textbf{0.95} & 0.78 & \textbf{0.95} & 0.47 & 0.57\\
    Welding    & 0.01 & 0.22& 0.02 & \textbf{0.37} & 0.01 & 0.01\\
        \midrule
    Photovoltaic & 0.85 & 0.53 & 0.97 & \textbf{1.00} & 0.68 & 0.64\\
     Module       & 0.11 & 0.01 & 0.74 & \textbf{0.85} & 0.03 & 0.05\\
        \midrule
    Wind     & 0.91 & 0.80 & \textbf{0.97} & 0.88 & 0.48 & 0.85\\
    Turbine  & 0.04 & 0.06 & \textbf{0.67} & 0.45 & 0.01 & 0.12\\
    \hline
    Catenary  & \textbf{0.96} & 0.93 & 0.95 & 0.95 & 0.61 & 0.81\\
    Dropper   & 0.22 & \textbf{0.60} & \textbf{0.60} & 0.48 & 0.05 & 0.07\\
        \midrule
    Nut and   & \textbf{0.89} & 0.67 & 0.55 & 0.84 & 0.50 & 0.52\\
    Bolt      & \textbf{0.20} & 0.05 & 0.01 & 0.02 & 0.01 & 0.01\\
        \midrule
    Witness & 0.57 & \textbf{0.71} & 0.51 & 0.52 & 0.49 & 0.59\\
    Mark    & \textbf{0.03} & 0.02 & 0.01 & 0.01 & 0.01 & 0.01\\
        \midrule
    Mean  & 0.82 & 0.73 & 0.75 & \textbf{0.85} & 0.53 & 0.64 \\
          & 0.09 & 0.14 & 0.29 & \textbf{0.35} & 0.02 & 0.04 \\
        \bottomrule
    \end{tabular}
  }
  \vspace{-5pt}
  \caption{Results of the evaluated methods when applied to the classification of anomalous images. For each method, the AUROC (top row) and Recall@1\%FPR (bottom row) are given. RD stands for Reverse Distillation and AE stands for Auto-Encoder.}
  \label{tab:exp_classification_results}
  \vspace{-5pt}
\end{table}

\begin{table}
  \centering
  \resizebox{\linewidth}{!}{
    \begin{tabular}{ccccccc}
        \toprule
    \textbf{Scenarios} &  \textbf{RD} & \textbf{PatchCore} & \textbf{FastFlow} & \textbf{DRAEM} & \textbf{AE} & \textbf{UniAD}\\
        \midrule
    Electrical & \multirow{2}{*}{0.90} & \multirow{2}{*}{0.64} & \multirow{2}{*}{0.74} & \multirow{2}{*}{\textbf{0.98}} & \multirow{2}{*}{0.51} & \multirow{2}{*}{0.64}\\
     Insulator &  &  & &  &  & \\
        \midrule
    Metal       & \multirow{2}{*}{0.68} & \multirow{2}{*}{0.47} & \multirow{2}{*}{\textbf{0.87}} & \multirow{2}{*}{0.64} & \multirow{2}{*}{0.60} & \multirow{2}{*}{0.76}\\
    Welding     &  &  &  &  &  & \\
        \midrule
    Photovoltaic & \multirow{2}{*}{0.92} & \multirow{2}{*}{0.52} & \multirow{2}{*}{0.59} & \multirow{2}{*}{\textbf{0.98}} & \multirow{2}{*}{0.85} & \multirow{2}{*}{0.69}\\
     Module       &  &  &  & & & \\
        \midrule
    Wind     & \multirow{2}{*}{0.95} & \multirow{2}{*}{0.73} & \multirow{2}{*}{\textbf{0.96}} & \multirow{2}{*}{0.92} & \multirow{2}{*}{0.45} & \multirow{2}{*}{0.89}\\
    Turbine  &  &  &  &  &  & \\
    \hline
    Catenary   & \multirow{2}{*}{0.76} & \multirow{2}{*}{0.86} & \multirow{2}{*}{\textbf{0.87}} & \multirow{2}{*}{0.53} & \multirow{2}{*}{0.59} & \multirow{2}{*}{\textbf{0.87}}\\
    Dropper    &  &  &  &  &  & \\
        \midrule
    Nut and    & \multirow{2}{*}{\textbf{0.94}} & \multirow{2}{*}{0.89} & \multirow{2}{*}{0.66} & \multirow{2}{*}{0.72} & \multirow{2}{*}{0.48} & \multirow{2}{*}{0.74}\\
    Bolt       &  &  &  &  &  & \\
        \midrule
    Witness & \multirow{2}{*}{0.10} & \multirow{2}{*}{0.65} & \multirow{2}{*}{0.45} & \multirow{2}{*}{0.48} & \multirow{2}{*}{0.56} & \multirow{2}{*}{\textbf{0.79}}\\
    Mark    &  &  &  &  &  & \\
        \midrule
    Mean  & 0.75 & 0.68 & 0.73 & 0.75 &  0.58 & \textbf{0.77}\\
        \bottomrule
    \end{tabular}
  }
  \vspace{-5pt}
  \caption{Results of the evaluated methods when applied to the segmentation of anomalous regions. For each method, the AUPRO is given. RD stands for Reverse Distillation and AE stands for Auto-Encoder.}
  \label{tab:exp_segmentation_results}
  \vspace{-15pt}
\end{table}

\subsection{Overall Results}

Evaluation results for the classification of anomalous images  are given for all methods and dataset categories in Tables \ref{tab:exp_classification_results}.
None of the evaluated methods emerges as a clear winner.
Considering only the mean AUROC, DERAM is the best and achives 0.85, while  the first four method achieve image-level AUROC scores of more than 0.98 on MVTec AD.
%
%
This verifies that outdoor inspection is more challenging than indoor inspection for all classes of algorithms due to uncontrolled environment.

From Tables~\ref{tab:exp_classification_results}, we can also observe that none of the methods manages to consistently perform well across all surface and logical anomaly.
DRAEM is more preferable for surface anomaly and Reverse Distillation is more preferable for logical anomaly.
In addition, the two metrics AUROC and Recall@1\%FPR does not satisfy the monotonic relation.
For example, Reverse Distillation achieve the highest AUROC 0.96 on the catenary dropper but not the highest Recall@1\%FPR.

Tables~\ref{tab:exp_classification_results} also shows that a higher AUROC is not always means a higher recall at 1\% FPR.
For example Reverse Distillation achieves highest AUROC with 0.96 on the wind turbine subset but obtains Recall@1\%FPR with 0.22 which is near three times smaller than than the highest score 0.60.
From the view of the metric Recall@1\%FPR, existing methods are far from applicable for outdoor inspection.

Evaluation results for the segmentation of anomalous regions are given for all methods and dataset categories in Tables~\ref{tab:exp_segmentation_results}.
We can observe that a high AUPRO does not necessarily
coincide with a high AUROC or a high Recall@1\%FPR.


\begin{figure*}[t]
    \vspace{-11pt}
    \centering
    \includegraphics[width=0.9 \textwidth]{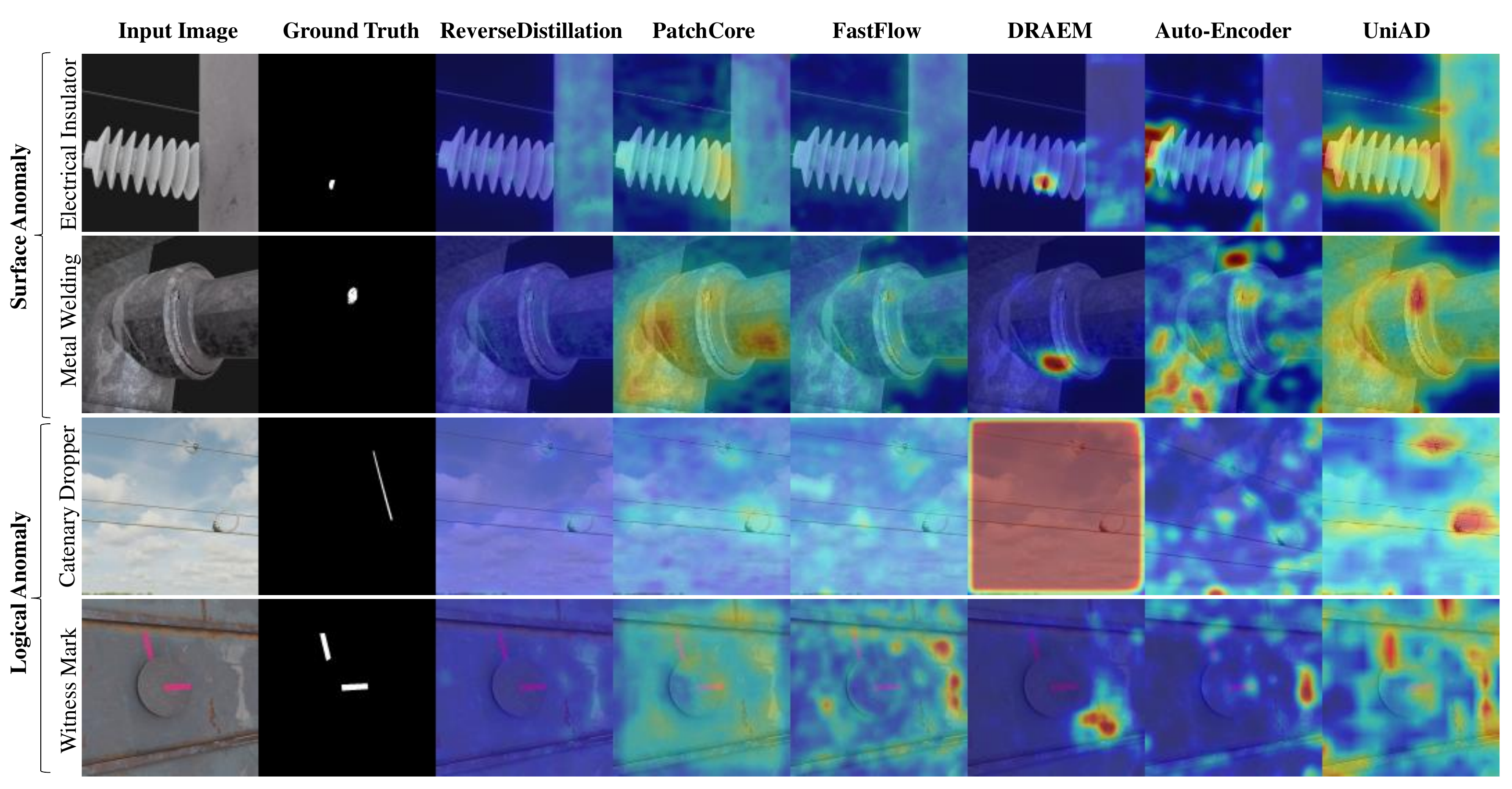}
    \caption{
    Qualitative anomaly segmentation results for each evaluated method on the MIAD dataset.}
    \label{fig:qualitative_results}
\end{figure*}

Qualitative results of each method are shown in Figure~\ref{fig:qualitative_results}, majority of results are failure cases due to uncontrolled viewpoints, background and surface.

Since the Auto-Encoder and UniAD performs poorly on MIAD, we will conduct further experiments only with the other four methods.  

\subsection{Impact of Outdoor Uncontrolled Factors}

%
In this section, we will study the impact of the three outdoor uncontrolled factors.
As described in Table~\ref{tab:statistic}, not all scenarios cover these three factors, so we select the electrical insulator and the witness mark as the representative of surface anomaly and logical anomaly, respectively.
We generate two baseline datasets under controlled environment, and individually 
 add UV, UB, US to the baseline. 
At the end, we add these three factors simultaneously, which is under the setting with our MIAD.
The results are shown in Table~\ref{tab:exp_uncontrolled_factors_ei} and \ref{tab:exp_uncontrolled_factors_wm}.
UV, UB and US will bring significant drops in performance separately to most methods except PatchCore on the witness mark.
The three factors seems to have equal impact from the view of AUROC.
However, from the view of Recall@1\%FPR, US plays a dominant role for the surface anomaly.
The combination of these factors results in greater drops without exception.

\begin{table}
  \centering
  \resizebox{\linewidth}{!}{
    \begin{tabular}{ccccccc}
        \toprule
   \textbf{Factors} &  \textbf{RD} & \textbf{PatchCore} & \textbf{FastFlow} & \textbf{DRAEM} & \textbf{Mean}\\
        \midrule
    baseline  & 0.85/0.53 & 0.80/0.40 &  0.83/0.44 & 0.90/0.44 & 0.85/0.45 \\
        \midrule
    +UV & 0.89/0.37 & 0.64/0.02 & 0.70/0.18 & 0.78/0.23 & 0.75/0.20 \\ 
        \midrule
    +UB & 0.73/0.17 & 0.71/0.29 & 0.71/0.41 & 0.81/0.19 & 0.74/0.27 \\ 
        \midrule
    +US & 0.81/0.05 & 0.76/0.07 & 0.79/0.16 & 0.89/0.21 & 0.81/0.12 \\ 
        \midrule
    +UV+US+UB & 0.68/0.01 & 0.55/0.01 & 0.54/0.01 & 0.81/0.24 & 0.65/0.07\\
        \bottomrule
    \end{tabular}
  }
  \vspace{-5pt}
  \caption{Impact of uncontrolled factors under outdoor environment on the electrical insulator.
  For each method, the image-level AUROC (the first number) and Recall@1\%FPR (the second number) are given. RD stands for Reverse Distillation and AE stands for Auto-Encoder.}
  \label{tab:exp_uncontrolled_factors_ei}
  \vspace{-5pt}
\end{table}
\begin{table}
  \centering
  \resizebox{\linewidth}{!}{
    \begin{tabular}{ccccccc}
        \toprule
   \textbf{Factors} &  \textbf{RD} & \textbf{PatchCore} & \textbf{FastFlow} & \textbf{DRAEM} & \textbf{Mean}\\
        \midrule
    baseline  & 0.95/0.42 & 1.00/1.00 & 1.00/1.00 & 1.00/1.00 & 0.99/0.86 \\
        \midrule
    +UV & 0.61/0.04 & 1.00/1.00 & 0.88/0.06 & 0.66/0.08 & 0.79/0.30 \\ 
        \midrule
    +UB & 0.57/0.03 & 0.93/0.39 & 0.92/0.28 & 0.86/0.27 & 0.82/0.24 \\ 
        \midrule
    +US & 0.64/0.04 & 1.00/0.99 & 0.99/0.65 & 0.47/0.02 & 0.78/0.43 \\ 
        \midrule
    +UV+US+UB & 0.57/0.03 & 0.71/0.02 & 0.51/0.01 & 0.52/0.01 & 0.58/0.02 \\
        \bottomrule
    \end{tabular}
  }
  \vspace{-5pt}
  \caption{Impact of uncontrolled factors under outdoor environment on the witness mark.
  For each method, the image-level AUROC (the first number) and Recall@1\%FPR (the second number) are given. RD stands for Reverse Distillation and AE stands for Auto-Encoder.}
  \label{tab:exp_uncontrolled_factors_wm}
  \vspace{-15pt}
\end{table}

\subsection{Impact of Training Dataset Size}

According to Table~\ref{tab:compare_datasets}, the training dataset size per category of existing datasests for unsupervised anomaly detection is not greater than 1200.
The limited training dataset size is partly due to costly collection, but the main reason is that hundreds of samples is enough to stands for the distribution of roughly aligned non-defective samples.
We believe an order of magnitude larger training data size is essential for outdoor anomaly detection to describe the complex non-defective objects. 
In order to verify this thought, we sample training data size from 1000 to 10000 on the electrical insulator which is a represent for surface anomaly and the witness mark which is a represent for logical anomaly.
As shown in Table~\ref{tab:exp_dataset_size}, Reverse Distillation enjoys the growth of the training data size, while PatchCore and FastFlow obtain a little benefit after the training dataset size is larger than 2000.
DRAEM, which is more preferable for surface anomaly, benefit a lot from the extensive training data on the electrical insulator but fail on the witness mark.
Furthermore, PatchCore is failed to run when training data size is larger than 2000 due to GPU memory limitation.
How to exploit the bonus of massive non-defective samples in the outdoors is an interesting research direction.

\begin{table}
  \centering
  \resizebox{\linewidth}{!}{
    \begin{tabular}{cccccc}
        \toprule
   \textbf{Size} & \textbf{RD} & \textbf{PatchCore} & \textbf{FastFlow} & \textbf{DRAEM} \\
        \midrule
   1000(10\%)  & 0.56/0.35 & 0.54/0.67 & 0.51/0.51  & 0.63/0.51 \\ 
        \midrule
   2000(20\%)  & 0.62/0.44 & 0.55/0.65 & 0.53/0.51 & 0.67/0.51 \\ 
        \midrule
   4000(40\%)  & 0.65/0.48 & -/- & 0.53/0.50 & 0.67/0.51 \\
        \midrule
   8000(80\%)  & 0.67/0.48 & -/- & 0.54/0.51 & 0.78/0.52 \\
        \midrule
   10000(100\%)  & 0.68/\textbf{0.57} & -/- & 0.53/0.51 & \textbf{0.81}/0.52 \\ 
        \bottomrule
    \end{tabular}
  }
  \vspace{-5pt}
  \caption{Results of different training dataset size. For each algorithm, the image-level AUROC on the electrical insulator and witness mark is given. Due to GPU memory limitation, PatchCore is failed to run when training data size is larger than 2000.}
  \label{tab:exp_dataset_size}
  \vspace{-15pt}
\end{table}

\section{Conclusion}
We introduce the MIAD dataset, a novel dataset for unsupervised anomaly detection in various maintenance inspections.
This dataset is intended for tasks under uncontrolled environments including uncontrolled viewpoints, uncontrolled backgrounds and uncontrolled surfaces.
It provides researchers with sufficient images, which are an order of magnitude larger than existing datasets, to facilitate algorithms to learn the complex distribution of outdoor non-defective samples.
Some representative methods including Reverse Distillation, PatchCore, FastFlow, DRAEM, Auto-Encoders and UniAD are evaluated on the MIAD dataset, and the significant drops in performance demonstrate that the maintenance inspection is more challenging than manufacturing inspection.
We expect that MIAD can attract more attention of the research community on the outdoor maintenance
inspection and that worthwhile future work can be spawned from the proposed dataset. 


\clearpage
\twocolumn[
\begin{@twocolumnfalse}
 \section*{\centering{Supplementary Material for \\ \emph{MIAD: A Maintenance Inspection Dataset for Unsupervised Anomaly Detection\\[25pt]}}}
\end{@twocolumnfalse}
]

In this supplementary material, we first describe the implementation details of each evaluated methods in Section~\ref{sec:A1}. Then we illustrate more qualitative examples on our MAID dataset in Section~\ref{sec:A2}.

\section{Implementation Details on Experiments}\label{sec:A1}


\subsection{Reverse Distillation}
We use the publicly available code in Anomalib~\cite{anomalib} as the implementation of Reverse Distillation~\cite{rd}.
%
Both training and testing images are zoomed to the input size of 256 × 256 pixels.
No data augmentation is applied, as this requires prior knowledge about class-retaining augmentations.
We adopt WideResNet50 as the backbone in the T-S model.
%
During training, we utilize Adam optimizer~\cite{kingma2014adam} with the learning rate of 0.005, $\beta_1$ of 0.5 and $\beta_2$ of 0.999.
All experiments on the MIAD dataset are run with a batch size of 32 on 1 GPU (NVIDIA Tesla V100 32GB).
The other parameters not mentioned are consistent with the default configuration in Anomalib.

\subsection{PatchCore}

We use the publicly available code in Anomalib~\cite{anomalib} as the implementation of PatchCore~\cite{patchcore}.
Both training and testing images are zoomed to the input size of 224 × 224 pixels and no data augmentation is applied.
We adopt WideResNet50 as the backbone.
All experiments on the MIAD dataset are run with a batch size of 32 on 1 GPU (NVIDIA Tesla V100 32GB).
Specifically, due to the limitation of GPU memory, PatchCore failed to run when training data size is larger than 2000.
Therefore, we randomly sample 2000 images to train the model.
The other parameters not mentioned are consistent with the default configuration in Anomalib.

\subsection{FastFlow}

We implement the code of FastFlow following the setting of the original paper~\cite{fastflow}. 
Both training and testing images are resized to the input size of 256 × 256 pixels and no data augmentation is applied.
We adopt Wide-ResNet50-2 as the backbone and directly use the embedding of the last layer in the first three blocks, and then put these features into
the 2D flow model to obtain their respective anomaly heatmaps, then we average the outputs of each heatmap. 
All the backbones used in the network are initialized from Imagenet pretrained weights. 
We use 8-step flows for Wide-ResNet50-2.  
We train this model using Adam optimizer with the learning rate of 1e-3 and weight
decay of 1e-5. 
We evaluate the final results after 500 epochs training with a batch size of 32 on 1 GPU (NVIDIA 1080Ti 12GB).
\subsection{DRAEM}

For the implementation of DRAEM~\cite{zavrtanik2021draem}, we use the publicly available code at \href{https://github.com/vitjanz/draem}{https://github.com/vitjanz/draem}.
Both training and testing images are zoomed to the input size of 256 $\times$ 256 pixels.
Random augmentation sampling is applied during training by a set of 3 random augmentation functions sampled from the
set: $\{$posterize, sharpness,solarize, equalize, brightness change, color change, auto-contrast$\}$.
Additional image rotation in the range of $(-45,45)$ degrees is used as a data augmentation method on non-defective images during training.
The neural network, which consists of a reconstructive sub-network and a discriminative sub-network following DRAEM~\cite{zavrtanik2021draem}, is trained for 700 epochs with a batch size of 8 on 1 GPU (NVIDIA Tesla V100 32GB).
The learning rate is set to $10^{-4}$ and is multiplied by 0.2 after 560 and 630 epochs with the Adam~\cite{kingma2014adam} optimizer. 
The $l_2$ and SSIM loss~\cite{ssim} are applied on the reconstructive sub-network, 
and the Focal Loss~\cite{focal_loss} is applied on the discriminative sub-network output to increase robustness towards accurate segmentation of hard examples.

\begin{figure*}[t]
    \centering
    \includegraphics[width=0.95 \textwidth]{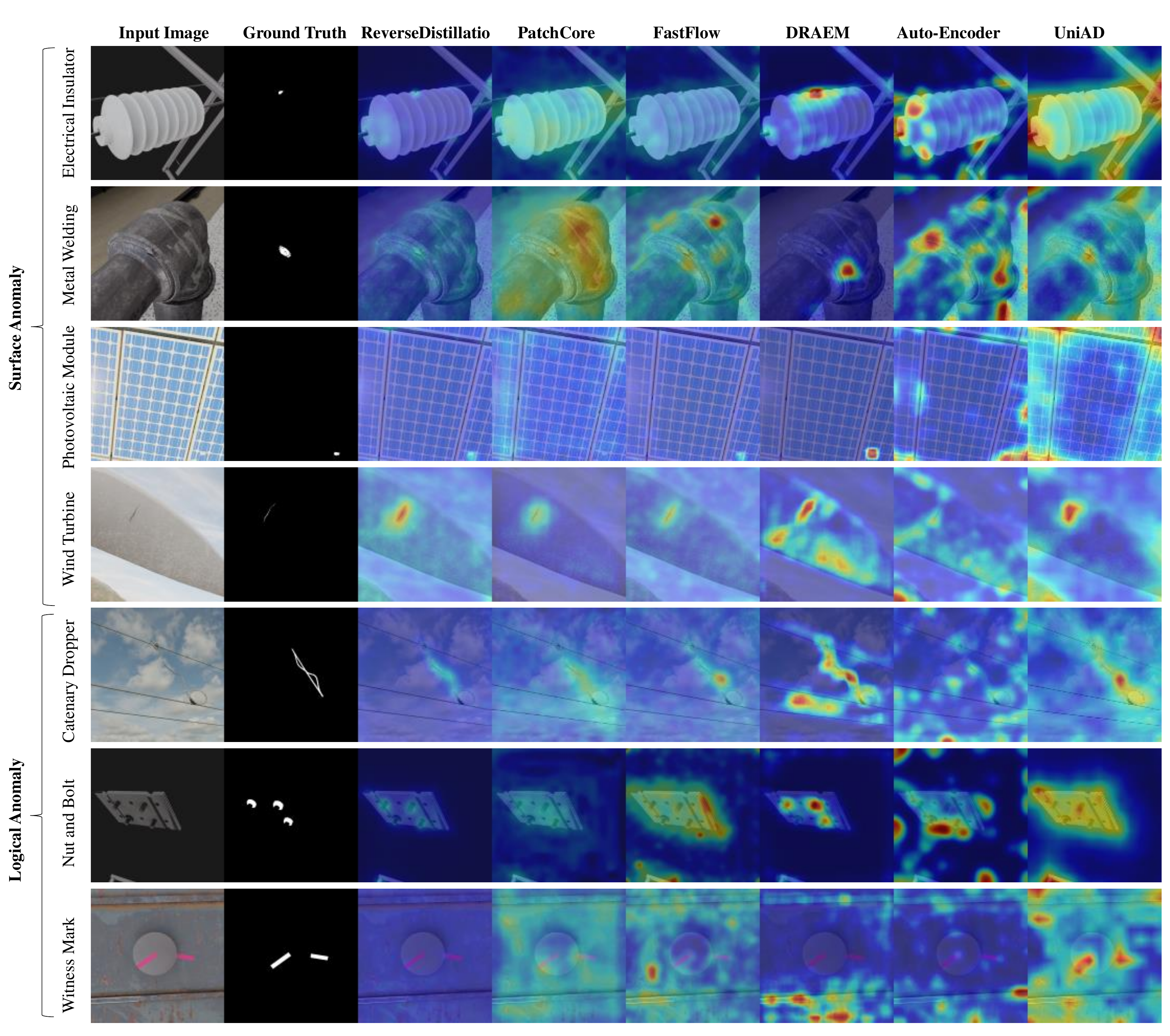}
    \caption{Qualitative anomaly segmentation results for each evaluated method on the MIAD dataset.}
    \label{fig:supp_qualitative_results}
\end{figure*}

\subsection{Auto-Encoder}

We implement the code of the Auto-Encoder following the setting of MVTecAD~\cite{mvtec_ad}.
A modified U-Net~\cite{ronneberger2015u} is used for the backbone instead of a simple stack of convolution layers.
All input RGB image is normalized by a mean of 0.5 and an standard deviation of 0.5.
Both training and testing images are zoomed to the input size of 256 $\times$ 256 pixels and no other data augmentation is used.
All experiments on the MIAD dataset are run for 100 epochs with a batch size of 16 on 1 GPU (NVIDIA Tesla V100 32GB).
All models are trained using the Adam~\cite{kingma2014adam} optimizer with a learning rate of 0.0001 and a weight decay of 0.0005.

\subsection{UniAD}
For the implementation of UniAD~\cite{uni_ad}, we use the publicly code at
\href{https://github.com/zhiyuanyou/UniAD}{https://github.com/zhiyuanyou/UniAD}.
A pre-trained EfficientNet-b4~\cite{tan2019efficientnet} model is used for feature extraction, and the feature maps from stage-1 to stage-4 are resized and concatenated together to form a 272-channel feature map.
Both training and testing images are zoomed to the input size of 224 $\times$ 224 pixels and no data augmentation is applied.
The model is trained for 500 epochs on 8 GPUs (NVIDIA Tesla V100 32GB) with batch size 64.
AdamW~\cite{loshchilov2017decoupled} optimizer with weight decay 0.0001 is used and the learning rate is 0.0001 initially, and dropped by 0.1 after 400 epochs.
Other settings are consistent with the original code.

\section{More Anomaly Localization Visualizations}\label{sec:A2}

More qualitative results of each method are shown in Figure~\ref{fig:supp_qualitative_results}.
The photovoltaic module scenario is only affected by uncontrolled viewpoints,
and is similar to indoor texture anomaly in MVTec AD.
Therefore, many methods including Reverse Distillation, FastFlow and DRAEM perform well.
The wind turbine is relatively easy for Reverse Distillation, PatchCore, FastFlow and DRAEM.
However, the other scenarios are very difficult. 
For example, methods fail on the electrical insulator, the nut and bolt mainly due to uncontrolled viewpoints.
Methods fail on the metal welding mainly result from uncontrolled surface.
Methods fail on the catenary droper mainly because of uncontrolled background.
And the combination of uncontrolled viewpoints, background and surface result in the failure cases of all methods on the witness mark.

Moreover, DRAEM is more preferable for surface anomaly because it is intended for surface anomaly detection and simulates surface anomalies by augmented images.
However logical anomalies can not be simulated well by simple data augmentation, which coincides with the imperfect performance of DRAEM.

{\small
\bibliographystyle{ieee_fullname}
\bibliography{ref}

\begin{thebibliography}{10}\itemsep=-1pt

\bibitem{anomalib}
Samet Akcay, Dick Ameln, Ashwin Vaidya, Barath Lakshmanan, Nilesh Ahuja, and
  Utku Genc.
\newblock Anomalib: A deep learning library for anomaly detection, 2022.

\bibitem{mvtec_loco}
Paul Bergmann, Kilian Batzner, Michael Fauser, David Sattlegger, and Carsten
  Steger.
\newblock Beyond dents and scratches: Logical constraints in unsupervised
  anomaly detection and localization.
\newblock {\em IJCV}, 2022.

\bibitem{mvtec_ad}
Paul Bergmann, Michael Fauser, David Sattlegger, and Carsten Steger.
\newblock {MVTec AD -- A comprehensive real-world dataset for unsupervised
  anomaly detection}.
\newblock In {\em CVPR}, 2019.

\bibitem{bergmann2021mvtec}
Paul Bergmann, Xin Jin, David Sattlegger, and Carsten Steger.
\newblock {The MVTec 3D-AD dataset for unsupervised 3D anomaly detection and
  localization}.
\newblock {\em arXiv preprint arXiv:2112.09045}, 2021.

\bibitem{ssim_ae}
Paul Bergmann, Sindy Löwe, Michael Fauser, David Sattlegger, and Carsten
  Steger.
\newblock Improving unsupervised defect segmentation by applying structural
  similarity to autoencoders.
\newblock In {\em 14th International Joint Conference on Computer Vision,
  Imaging and Computer Graphics Theory and Applications}, 2019.

\bibitem{review_food}
Tadhg Brosnan and Da-Wen Sun.
\newblock Improving quality inspection of food products by computer
  vision––a review.
\newblock {\em Journal of food engineering}, 2004.

\bibitem{artical_catenary}
Junwen Chen, Zhigang Liu, Hongrui Wang, Alfredo Núñez, and Zhiwei Han.
\newblock Automatic defect detection of fasteners on the catenary support
  device using deep convolutional neural network.
\newblock {\em IEEE Transactions on Instrumentation and Measurement}, 2017.

\bibitem{spade}
Niv Cohen and Yedid Hoshen.
\newblock Sub-image anomaly detection with deep pyramid correspondences.
\newblock In {\em arXiv preprint arXiv:2005.02357}, 2020.

\bibitem{survey_uad}
Yajie Cui, Zhaoxiang Liu, and Shiguo Lian.
\newblock A survey on unsupervised industrial anomaly detection algorithms.
\newblock In {\em arXiv preprint arXiv:2204.11161}, 2022.

\bibitem{padim}
Thomas Defard, Aleksandr Setkov, Angelique Loesch, and Romaric Audigier.
\newblock {PaDiM: a patch distribution modeling framework for anomaly detection
  and localization}.
\newblock In {\em ICPR}, 2021.

\bibitem{Deitsch2019elpv}
Sergiu Deitsch, Vincent Christlein, Stephan Berger, Claudia Buerhop-Lutz,
  Andreas Maier, Florian Gallwitz, and Christian Riess.
\newblock Automatic classification of defective photovoltaic module cells in
  electroluminescence images.
\newblock {\em Solar Energy}, 185:455--468, June 2019.

\bibitem{rd}
Hanqiu Deng and Xingyu Li.
\newblock Anomaly detection via reverse distillation from one-class embedding.
\newblock In {\em CVPR}, 2022.

\bibitem{denninger2019blenderproc}
Maximilian Denninger, Martin Sundermeyer, Dominik Winkelbauer, Youssef Zidan,
  Dmitry Olefir, Mohamad Elbadrawy, Ahsan Lodhi, and Harinandan Katam.
\newblock Blenderproc.
\newblock {\em arXiv preprint arXiv:1911.01911}, 2019.

\bibitem{review_aoi}
Abd Al Rahman M.~Abu Ebayyeh and Alireza Mousavi.
\newblock A review and analysis of automatic optical inspection and quality
  monitoring methods in electronics industry.
\newblock {\em IEEE Access}, 2020.

\bibitem{gan}
Ian Goodfellow, Jean Pouget-Abadie, Mehdi Mirza, Bing Xu, David Warde-Farley,
  Sherjil Ozair, Aaron Courville, and Yoshua Bengio.
\newblock Generative adversarial networks.
\newblock {\em Communications of the ACM}, 2020.

\bibitem{cflow}
Denis Gudovskiy, Shun Ishizaka, and Kazuki Kozuka.
\newblock {CFLOW-AD: Real-time unsupervised anomaly detection with localization
  via conditional normalizing flows}.
\newblock {\em arXiv preprint arXiv:2107.12571}, 2021.

\bibitem{huang2020surface}
Yibin Huang, Congying Qiu, and Kui Yuan.
\newblock Surface defect saliency of magnetic tile.
\newblock {\em The Visual Computer}, 36(1):85--96, 2020.

\bibitem{kingma2014adam}
Diederik~P Kingma and Jimmy Ba.
\newblock Adam: A method for stochastic optimization.
\newblock {\em arXiv preprint arXiv:1412.6980}, 2014.

\bibitem{vae}
Diederik~P Kingma and Max Welling.
\newblock Auto-encoding variational bayes.
\newblock In {\em ICLR}, 2014.

\bibitem{lecun1989}
Yann LeCun.
\newblock Generalization and network design strategies.
\newblock {\em Connectionism in perspective}, 1989.

\bibitem{cfa}
Sungwook Lee, Seunghyun Lee, and Byung~Cheol Song.
\newblock Cfa: Coupled-hypersphere-based feature adaptation for target-oriented
  anomaly localization.
\newblock {\em arXiv preprint arXiv:2206.04325}, 2022.

\bibitem{li2021cutpaste}
Chun-Liang Li, Kihyuk Sohn, Jinsung Yoon, and Tomas Pfister.
\newblock Cutpaste: Self-supervised learning for anomaly detection and
  localization.
\newblock In {\em Proceedings of the IEEE/CVF Conference on Computer Vision and
  Pattern Recognition}, pages 9664--9674, 2021.

\bibitem{artical_photovoltaic}
Xiaoxia Li, Qiang Yang, Zhebo Chen, and Wenjun Yan.
\newblock Visible defects detection based on uav‐based inspection in
  large‐scale photovoltaic systems.
\newblock {\em IET Renewable Power Generation}, 2017.

\bibitem{focal_loss}
Tsung-Yi Lin, Priya Goyal, Ross Girshick, Kaiming He, and Piotr Dollár.
\newblock Focal loss for dense object detection.
\newblock In {\em ICCV}, 2017.

\bibitem{loshchilov2017decoupled}
Ilya Loshchilov and Frank Hutter.
\newblock Decoupled weight decay regularization.
\newblock {\em arXiv preprint arXiv:1711.05101}, 2017.

\bibitem{mishra2021vt}
Pankaj Mishra, Riccardo Verk, Daniele Fornasier, Claudio Piciarelli, and
  Gian~Luca Foresti.
\newblock {VT-ADL: A vision transformer network for image anomaly detection and
  localization}.
\newblock In {\em 2021 IEEE 30th International Symposium on Industrial
  Electronics (ISIE)}, pages 01--06. IEEE, 2021.

\bibitem{PBR}
Matt Pharr, Wenzel Jakob, and Greg Humphreys.
\newblock {\em Physically based rendering: From theory to implementation}.
\newblock Morgan Kaufmann, 2016.

\bibitem{nflow}
Danilo Rezende and Shakir Mohamed.
\newblock Variational inference with normalizing flows.
\newblock In {\em PMLR}, 2015.

\bibitem{ronneberger2015u}
Olaf Ronneberger, Philipp Fischer, and Thomas Brox.
\newblock U-net: Convolutional networks for biomedical image segmentation.
\newblock In {\em International Conference on Medical image computing and
  computer-assisted intervention}, pages 234--241. Springer, 2015.

\bibitem{patchcore}
Karsten Roth, Latha Pemula, Joaquin Zepeda, Bernhard Schölkopf, Thomas Brox,
  and Peter Gehler.
\newblock Towards total recall in industrial anomaly detection.
\newblock In {\em CVPR}, 2022.

\bibitem{differnet}
Marco Rudolph, Bastian Wandt, and Bodo Rosenha.
\newblock Same same but differnet: Semi-supervised defect detection with
  normalizing flows.
\newblock In {\em Proceedings of the IEEE/CVF winter conference on applications
  of computer vision}, 2021.

\bibitem{structuring_ae}
Marco Rudolph, Bastian Wandt, and Bodo Rosenhahn.
\newblock Structuring autoencoders.
\newblock In {\em ICCV}, 2019.

\bibitem{aitex}
Javier Silvestre-Blanes, Teresa Albero-Albero, Ignacio Miralles, Rubén
  Pérez-Llorens, and Jorge Moreno.
\newblock A public fabric database for defect detection methods and results.
\newblock {\em Autex Research Journal}, 2019.

\bibitem{review_power_transmission}
A. Sriram and T.~D. Sudhakar.
\newblock Technology revolution in the inspection of power transmission lines-a
  literature review.
\newblock In {\em 2021 7th International Conference on Electrical Energy
  Systems (ICEES)}, pages 256--262. IEEE, 2021.

\bibitem{Tabernik2019JIM}
Domen Tabernik, Samo {\v{S}}ela, Jure Skvar{\v{c}}, and Danijel Sko{\v{c}}aj.
\newblock Segmentation-based deep-learning approach for surface-defect
  detection.
\newblock {\em Journal of Intelligent Manufacturing}, May 2019.

\bibitem{tan2019efficientnet}
Mingxing Tan and Quoc Le.
\newblock Efficientnet: Rethinking model scaling for convolutional neural
  networks.
\newblock In {\em International conference on machine learning}, pages
  6105--6114. PMLR, 2019.

\bibitem{artical_metallic_surface}
Xian Tao, Dapeng Zhang, Wenzhi Ma, Xilong Liu, and De Xu.
\newblock Automatic metallic surface defect detection and recognition with
  convolutional neural networks.
\newblock {\em Applied Sciences}, 2018.

\bibitem{tao2018detection}
Xian Tao, Dapeng Zhang, Zihao Wang, Xilong Liu, Hongyan Zhang, and De Xu.
\newblock Detection of power line insulator defects using aerial images
  analyzed with convolutional neural networks.
\newblock {\em IEEE Transactions on Systems, Man, and Cybernetics: Systems},
  50(4):1486--1498, 2018.

\bibitem{stpm}
Guodong Wang, Shumin Han, Errui Ding, and Di Huang.
\newblock Student-teacher feature pyramid matching for unsupervised anomaly
  detection.
\newblock In {\em BMVC}, 2021.

\bibitem{review_wind_turbine}
Wenjie Wang, Yu Xue, Chengkuan He, and Yongnian Zhao.
\newblock Review of the typical damage and damage-detection methods of large
  wind turbine blades.
\newblock {\em Energies}, 2022.

\bibitem{ssim}
Zhou Wang, Alan~C. Bovik, Hamid~R. Sheikh, and Eero~P. Simoncelli.
\newblock Image quality assessment: from error visibility to structural
  similarity.
\newblock {\em IEEE Transactions on image processing}, 2004.

\bibitem{dagm}
Matthias Wieler and Tobias Hahn.
\newblock Weakly supervised learning for industrial optical inspection.
\newblock In {\em DAGM Symposium}, 2007.

\bibitem{rstpm}
Shinji Yamada and Kazuhiro Hotta.
\newblock Reconstruction student with attention for student-teacher pyramid
  matching.
\newblock In {\em arXiv preprint arXiv:2111.15376}, 2021.

\bibitem{uni_ad}
Zhiyuan You, Lei Cui, Yujun Shen, Kai Yang, Xin Lu, Yu Zheng, and Xinyi Le.
\newblock A unified model for multi-class anomaly detection.
\newblock In {\em NeurIPS}, 2022.

\bibitem{fastflow}
Jiawei Yu, Ye Zheng, Xiang Wang, Wei Li, Yushuang Wu, Rui Zhao, and Liwei Wu.
\newblock {FastFlow: Unsupervised anomaly detection and localization via 2D
  normalizing flows}.
\newblock {\em arXiv preprint arXiv:2111.07677}, 2021.

\bibitem{zavrtanik2021draem}
Vitjan Zavrtanik, Matej Kristan, and Danijel Sko{\v{c}}aj.
\newblock {DRAEM -- A discriminatively trained reconstruction embedding for
  surface anomaly detection}.
\newblock In {\em Proceedings of the IEEE/CVF International Conference on
  Computer Vision}, pages 8330--8339, 2021.

\bibitem{zheng2022focus}
Ye Zheng, Xiang Wang, Rui Deng, Tianpeng Bao, Rui Zhao, and Liwei Wu.
\newblock Focus your distribution: Coarse-to-fine non-contrastive learning for
  anomaly detection and localization.
\newblock In {\em 2022 IEEE International Conference on Multimedia and Expo
  (ICME)}, pages 1--6. IEEE, 2022.

\end{thebibliography}
}

\end{document}